%% file: exponpc.tex
\documentclass[letterpaper, 10 pt, journal, twoside]{ieeetran}  

\IEEEoverridecommandlockouts                              

\IEEEpubid{\begin{minipage}{\textwidth}\ \\[12pt] 
\end{minipage}}

\usepackage{times}
\usepackage[pdftex]{graphicx}
\usepackage{subfigure}
\usepackage{capt-of}
\usepackage{amsmath,amssymb,amsopn,amstext,amsfonts}
\usepackage{cancel}
\usepackage[space]{cite}
\usepackage{pdfsync}
\usepackage{balance}
\usepackage{color}
\usepackage{mathtools}
\usepackage{bm}

\usepackage{diagbox}
\usepackage{float}
\usepackage{epstopdf}
\usepackage{pifont}
\usepackage{fixltx2e}
\usepackage{amsmath}
\usepackage{multirow}
\usepackage{url}
\usepackage{verbatim}
\usepackage[linkcolor=black,citecolor=black,urlcolor=black,colorlinks=true]{hyperref}
\usepackage{soul,xcolor}
\usepackage[linesnumbered,ruled,vlined]{algorithm2e}
\usepackage{subfiles}
\usepackage{array}
\usepackage{hhline}
\usepackage{makecell}
\usepackage{booktabs}
\usepackage{diagbox}
\usepackage{threeparttable} 
\graphicspath{{./figures/}{../figures/}}
\DeclareGraphicsExtensions{.png,.jpg,.eps,.pdf,.jpeg}
\makeatletter
\let\@oldmaketitle\@maketitle
\renewcommand{\@maketitle}{\@oldmaketitle
    \vspace{0.4cm}
    \centering  
    \setcounter{figure}{0}
    \includegraphics[width=2.0\columnwidth]{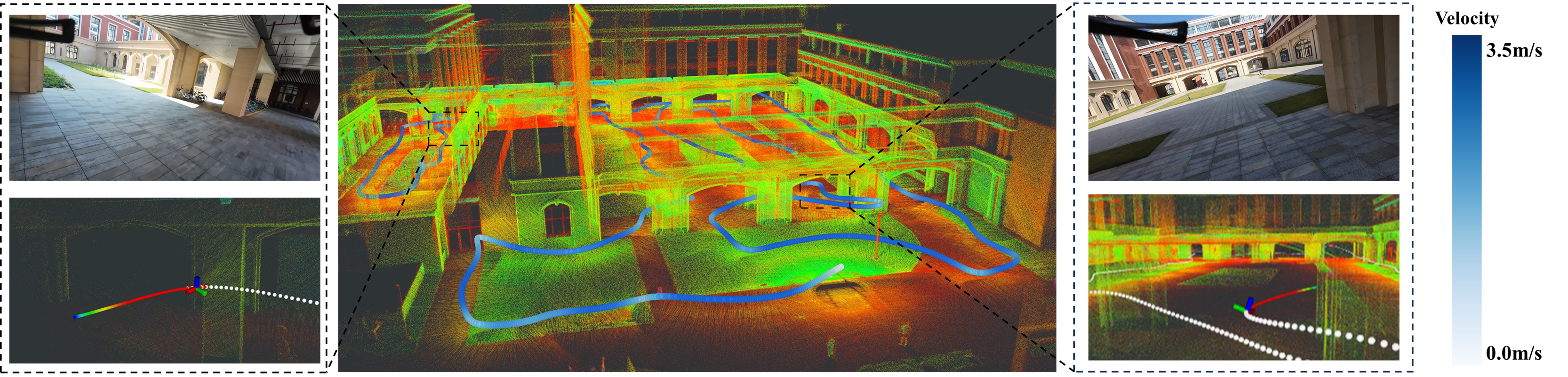}
   
   \label{fig:building}
   \vspace{-0.2cm}
    \captionof{figure}{
     Autonomous exploration of a large-scale scene with complex structures, spanning thousands of square meters. The central image displays the exploration result, showing the online-generated point cloud map and the UAV's flight trajectory. The trajectory color intensity indicates the UAV's velocity, with darker hues representing higher speeds. The side images illustrate two snapshots during the exploration process: the UAV's first-person view, current pose, and locally planned trajectories. The color of the local trajectories also denotes velocity, ranging from red (maximum speed) to blue (stationary).}
    \vspace{-0.9cm}
}
\makeatother
\title{
  EPIC: A Lightweight LiDAR-Based UAV Exploration Framework for Large-Scale Scenarios
  \vspace{-0.2cm}
}

\author{ Shuang Geng $^{1,2, * \,}$, Zelin Ning $^{1, * \,}$, Fu Zhang $^{2 \,}$, and Boyu Zhou $^{1, \dag}$

\thanks{\scriptsize{\textbf{$^{*}$Equal Contribution}, \textbf{$^{\dag}$Corresponding Author}. $^{1}$School of Artificial Intelligence, Sun Yat-Sen University, Zhuhai, China.  $^{2}$Department of Mechanical Engineering, University of Hong Kong, Hong Kong, China.}}

\thanks{\scriptsize{Email: shuangge@connect.hku.hk, ningzlin@mail2.sysu.edu.cn,}}\thanks{\scriptsize{\qquad \quad fuzhang@hku.hk, zhouby23@mail.sysu.edu.cn}}
\thanks{ \scriptsize{This work was completed during Geng's exchange period at Sun Yat-Sen University.}}
\thanks{\scriptsize{$^{\divideontimes}$To be released at \url{https://github.com/SYSU-STAR/EPIC}}}
\vspace{-0.5cm}   
}

\begin{document}
\maketitle

\vspace{0.2cm}
\begin{abstract}
Autonomous exploration is a fundamental problem for various applications of unmanned aerial vehicles (UAVs).
Recently, LiDAR-based exploration has gained significant attention due to its ability to generate high-precision point cloud maps of large-scale environments. While the point clouds are inherently informative for navigation, many existing exploration methods still rely on additional, often expensive, environmental representations. This reliance stems from two main reasons: the need for frontier detection or information gain computation, which typically depends on memory-intensive occupancy grid maps, and the high computational complexity of path planning directly on point clouds, primarily due to costly collision checking.
To address these limitations, we present EPIC, a lightweight LiDAR-based UAV exploration framework that directly exploits point cloud data to explore large-scale environments. 
EPIC introduces a novel observation map based on the quality of point clouds, treating the environment as a collection of small surface patches and evaluating their observation quality. It maintains and updates this quality using spatial hashing. By guiding the UAV from well-observed to poorly-observed areas, EPIC eliminates the need for global occupancy grid maps, while ensuring robust exploration and effective performance across diverse environments.
We also propose an incremental topological graph construction method operating directly on point clouds, enabling real-time path planning in large-scale environments.
Leveraging these components, we build a hierarchical planning framework that generates agile and energy-efficient trajectories, achieving significantly reduced memory consumption and computation time compared to most existing methods.
Extensive simulations and real-world experiments demonstrate that EPIC achieves faster exploration while significantly reducing memory consumption compared to state-of-the-art methods.
Our method will be made open source to benefit the community$^{\divideontimes}$.
 
\end{abstract}









\subfile{sections/intro.tex}

\subfile{sections/related.tex}

\subfile{sections/problem.tex}

\subfile{sections/frontier.tex}

\subfile{sections/planning.tex}

\subfile{sections/results.tex}

\vspace{-0.1cm}
\section{Conclusions}
In this letter, we present EPIC, a lightweight LiDAR-based UAV exploration framework for large-scale environments. It uses an observation-based environment representation directly derived from point clouds, eliminating memory-intensive global occupancy grid maps while preserving sufficient information for comprehensive exploration. 
We also develop an incremental topological graph construction method operating directly on point clouds, enabling efficient path planning.
Leveraging these components, we build a hierarchical planning framework, achieving significantly reduced memory consumption and computation time compared to existing methods.

\vspace{-0.3cm}
\addtolength{\textheight}{0.cm}   

\newlength{\bibitemsep}\setlength{\bibitemsep}{0.0\baselineskip}
\newlength{\bibparskip}\setlength{\bibparskip}{0.0pt}
\let\oldthebibliography\thebibliography
\renewcommand\thebibliography[1]{%
\oldthebibliography{#1}%
\setlength{\parskip}{\bibitemsep}%
\setlength{\itemsep}{\bibparskip}%
}

\bibliography{exponpc} 

\end{document}

%% file: sections/intro.tex
\section{Introduction}

\IEEEPARstart{i}{n} recent years, autonomous exploration algorithms have been widely applied in industrial inspection, mineral exploration, and search and rescue tasks.
While exploration tasks for indoor or small-scale scenarios have been well addressed\cite{zhou2021fuel, zhang2024falcon}, large-scale environment exploration poses significant challenges to limited onboard resources due to high memory consumption and computational overhead.

Recently, LiDAR-based exploration has garnered increasing attention, where high-precision point cloud maps of large-scale environments are incrementally constructed during exploration\cite{yang2021graph, gao2022meeting, tang2023bubble}. Although the online-generated point cloud maps are informative enough for navigation, many existing exploration methods still rely on additional, often expensive, environmental representations \cite{zhou2021fuel, zhang2024falcon, tang2023bubble, duberg2022ufoexplorer}. A key reason for this reliance is the need for frontier detection or information gain computation, which is a crucial step in determining the UAV's next destination. This process typically depends on an additional global occupancy grid map or its variants, such as UFOMap \cite{duberg2020ufomap}, which are memory-intensive in large-scale scenarios. Some methods employ geometric representations of the environment, such as convex polyhedrons \cite{yang2021graph} or star-convex polyhedrons \cite{gao2022meeting}. While offering benefits in terms of memory efficiency, they often fail to capture fine-grained environmental details, potentially leading to incomplete exploration.



Another challenge driving the reliance on expensive representations is the computational complexity of performing path planning directly on point clouds, primarily due to the high cost of collision checking. The need for frequent path searches in large-scale environments further exacerbates this issue. To address this, topological graphs can be constructed to accelerate path planning in expansive scenes. However, most methods for generating topological graphs are designed to operate on occupancy grid maps \cite{zhang2024falcon, musil2022spheremap}. While some approaches have been adapted for point cloud maps \cite{chen2022fast, guo2022dynamic}, they generally lack the capability for real-time, incremental construction in complex scenes, which is crucial for efficient exploration.

In this paper, we propose EPIC (\textbf{E}xploring on \textbf{P}o\textbf{I}nt \textbf{C}louds), a lightweight LiDAR-based UAV exploration framework that directly exploits point cloud data to explore large-scale environments. Our key idea is to maintain a compact yet informative representation, termed the observation map, which is derived directly from the quality of point clouds. This eliminates the need for a global occupancy grid map while retaining sufficient information for comprehensive exploration. To enable real-time path planning, we introduce an incremental topological graph construction method that operates directly on point clouds. 
Leveraging these components, we build a hierarchical planning framework that generates safe, smooth, and energy-efficient trajectories for the UAV, achieving significantly reduced and stable computation time per planning cycle compared to most existing methods.
We evaluated EPIC in both simulated and real-world environments, comparing it with state-of-the-art exploration methods. Experimental results demonstrate that our framework achieves faster exploration while significantly reducing memory consumption. The main contributions of this work are:

{1) A memory-efficient observation map that represents surface observation quality using spatial hashing, enabling UAV exploring from well-observed areas to poorly-observed areas without maintaining global occupancy maps,  significantly reduces memory consumption.}

{2) An incremental approach to constructing a topological graph by directly assessing spatial connectivity on point clouds, avoiding memory-intensive grid maps. This approach enables more efficient path searching through a sparser connectivity representation, thereby reducing the computational cost of global planning.}

\label{sec:intro}

%% file: sections/related.tex
\section{Related Work}
\label{sec:related}
\subsection{Environment Representation for Exploration}
\label{subs:related_env}

Environment representation is a fundamental component in UAV autonomous exploration, providing critical information to guide UAV to explore unknown spaces. Traditional frontier-based exploration methods \cite{zhou2021fuel, cao2021tare, tang2023bubble} typically rely on a global uniform occupancy grid map or its variants, such as UFOMap \cite{duberg2020ufomap}, where each grid cell is labeled to unknown, free, or occupied. These methods define the boundary between unknown and free areas as frontiers and navigate the UAV to observe them. {To avoid storing large amounts of data related to free and unknown space, Huang $et~al.$\cite{huang2024efficient}  proposed FAOmap (Frontier And Obstacle map), which only maintains information about frontiers and obstacle surfaces.} While such methods perform well in confined environments such as indoors or caves, they generate numerous unnecessary trajectories when applied to large open  scenes like the campus demonstrated in our experiments. This inefficiency is due to the UAV being directed to observe empty areas, such as the sky. 
Some newly proposed methods define the intersection of unknown, free, and occupied areas as surface frontiers\cite{zhang2024soar, surfaceNBVP}, guiding UAV exploration by focusing on the scene’s surface and thereby avoiding unnecessary trajectories. Although both approaches ensure comprehensive exploration of the environment, the requirement for occupancy grid maps results in high memory consumption, challenging the limited onboard resources.

Some other methods employ geometric methods to represent frontiers.
For example, Yang $et~al.$\cite{yang2021graph} utilize convex polyhedrons to represent detected regions and define distant polyhedrons as frontiers. 
This approach guides the UAV towards these frontiers and updating them accordingly. Gao $et~al.$\cite{gao2022meeting} employ a star-convex representation of detected areas, where triangular facets on the surface of star-convex polyhedrons are defined as frontiers. As the UAV navigates in the environment, this method generates new star-convex polyhedrons and eliminates triangular facets whose center points have been observed.
While these geometric approaches offer significant advantages in terms of memory utilization, simplified geometric models may not capture fine-grained environmental features and their coarse frontier update strategies potentially result in incomplete coverage of the environment.

Our method, termed the observation map, focuses on surfaces of the environment, avoiding the overhead of maintaining a global occupancy grid map and reducing unnecessary trajectories, while preserving sufficient information for comprehensive exploration of large-scale environments.

\subsection{Topological Graph Construction}
\label{subs:related_topograph}
Path searching is crucial for real-time exploration, as it is frequently executed in global guidance path planning. However, naive searching on occupancy grid maps or point cloud maps may be infeasible in real-time for long distances. To address this problem, several methods construct topological graphs for efficient path searching in large-scale environments.

Musil $et~al.$ \cite{musil2022spheremap} use an octree map to represent free regions, covering these areas with collision-free spheres and incrementally updating the topological map based on sphere connectivity. Zhang $et~al.$ \cite{zhang2024falcon} represent the exploration space using a uniform occupancy grid map and incrementally decompose the space into disjoint free and unknown zones based on voxel connectivity from the latest occupancy information, building a connectivity graph where zone centers act as vertices and zone connectivity defines the edges. Both approaches achieve incremental updates of topological graphs but require maintaining additional occupancy grid maps or octree maps, resulting in high memory consumption in large-scale environments. Methods like \cite{chen2022fast, guo2022dynamic} can construct topological graphs directly on point clouds but lack incremental construction capabilities. 

Our method is inspired by Zhang $et~al.$ \cite{zhang2024falcon}, decomposing the space into disjoint zones and incrementally constructing the topological graph based on their connectivity. However, unlike their approach, we construct directly on point clouds, significantly reducing memory consumption.

%% file: sections/problem.tex
\section{Problem Definition and System Overview}
\label{sec:problem}
The problem addressed in this letter is exploring a large-scale, unknown, bounded 3D space $H\in\mathbb{R}^3$ using an {unmanned aerial vehicle} (UAV) to create a dense point cloud map. 

The proposed exploration framework is consisted of an observation-based environment representation for exploration (\ref{sec:frontier}) and a real-time hierarchical planning framework (\ref{sec:path}). Upon receiving a new LiDAR point cloud scan, the observation map $\mathcal{M}_{obs}$, which records the quality of surface observations, is incrementally updated (\ref{subs:labelmap}). Based on $\mathcal{M}_{obs}$, frontiers $S_{ftr}$ are detected and clustered (\ref{subs:detection}).
To facilitate efficient path planning in large-scale environments, a topological graph $G$ is incrementally constructed on point clouds (\ref{subs:topo_graph}). Viewpoints are strategically selected based on topological reachability. Subsequently, a global guidance path starting from the current position and passing through all
viewpoints is calculated (\ref{subsec:global_planning}). Based on the global tour, energy-efficient and agile local trajectories are generated, guiding the UAV to visit these viewpoints sequentially (\ref{subs:traj}).
The exploration is considered finished if no frontier exists.

%% file: sections/frontier.tex
\section{Observation-based Exploration Environment Representation}
\label{sec:frontier}
In this section, we propose a novel observation-based environment representation method for UAV exploration, which comprises two key components: observation map $\mathcal{M}_{obs}$ and frontier cluster list $\mathcal{L}_{clusters}$.

$\mathcal{M}_{obs}$ records the quality of surface observations, which is determined by the LiDAR's observation distance and view direction. Based on the quality evaluation, observed surfaces $S_{obs}$ are labeled as either well-observed or poorly-observed.
The frontiers $S_{frt}$, defined as the boundary between well-observed surfaces $S_{well}$ and poorly-observed surfaces $S_{poorly}$, are clustered and stored in $\mathcal{L}_{clusters}$. When a new frame of point cloud is received, both $\mathcal{M}_{obs}$ and $\mathcal{L}_{clusters}$ are updated incrementally.




\subsection{Observation Map Construction}
\label{subs:labelmap}
To facilitate efficient query of surface observation quality, and avoid excessive memory consumption, we employ a spatial hashing map\cite{niessner2013real} to store the quality of surface observations. Defining it as observation map $\mathcal{M}_{obs}$, each element inside is a voxel representing a small surface patch and is labeled as well-observed or poorly-observed based on observation quality.
This selective storage, unlike memory-intensive occupancy grid maps, ensures that $\mathcal{M}_{obs}$ only focuses on the areas around the surface, thereby reducing memory consumption while retaining sufficient information for frontier detection. 

The observation quality of each surface patch is determined by distance constraint and view direction constraint. Consider a surface patch centered at $p_{s}$ and the LiDAR is located at $p_l$, the distance constraint is considered satisfied if 
\begin{equation}
\| p_l - p_s \|_2 \leq D,
\label{eq:criteria1}
\end{equation}
where $D$ is a predefined constant that ensures the LiDAR is close enough to the surface patch to obtain high-quality observations.

\begin{figure}[!t]
    \centering 
    \includegraphics[width=0.95\linewidth]{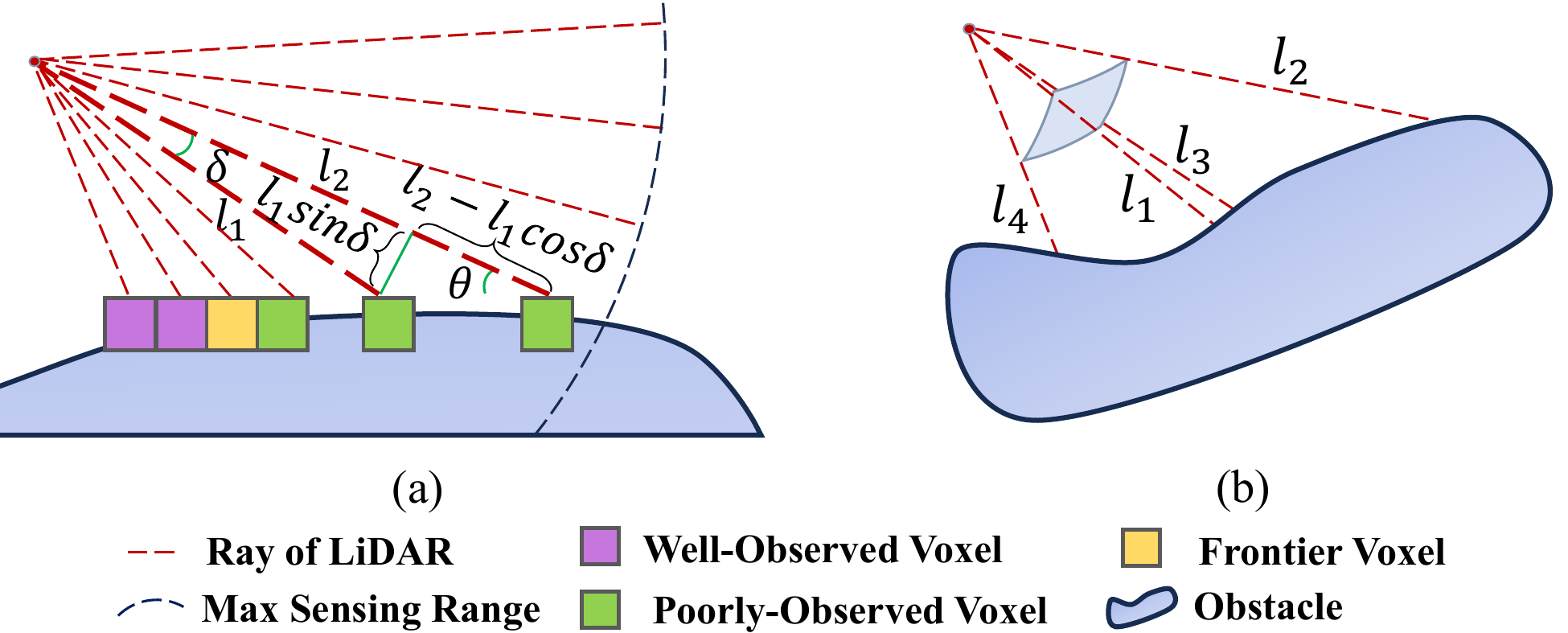}
    \caption{
   Evaluation of view direction quality. (a): A 2D example illustrating how adjacent LiDAR rays are used to evaluate view direction quality. (b): A pyramidal-shaped volume is formed by four adjacent LiDAR rays, and the view direction quality of a voxel, whose center falls within this volume, is determined by these four LiDAR rays.}
    \label{fig:pry}
    \vspace{-0.7cm}
\end{figure}

Given that the quality of observation improves as the LiDAR beam becomes more perpendicular to the surface patch, as illustrated in Fig. \ref{fig:pry}(a), for a 2D case, the quality of view direction is negatively correlated with $|90^\circ - \theta|$.  Assuming $l_2 \geq l_1$, $\theta$ can be can be computed by:

\begin{equation}
cot\theta = \frac{l_2-l_1 cos\delta}{l_1 sin\delta} = \frac{1}{sin\delta}\frac{l_2}{l_1}-cot\delta,
\label{eq:cot}
\end{equation}
where $\delta$ represents the constant angular separation between two adjacent LiDAR beams, a value that is determined by the mechanical configuration of the LiDAR.
According to equation~(\ref{eq:cot}), $|90^\circ - \theta|$ and $l_1 / l_2$ are negatively correlated. Therefore, we can use $l_1 / l_2$ to evaluate the quality of the view direction. We consider the view direction constraint satisfied when $l_1 / l_2$ exceeds a threshold $T$.

For 3D cases, as shown in Fig. \ref{fig:pry}(b), the sensing area of LiDAR is composed of a number of small pyramidal-shaped volumes. 
Consider a surface patch centered at $p_{i}$, falling within a pyramidal-shaped volume $f_{i}$. If the four rays composing $f_{i}$ mutually satisfy the aforementioned condition, we consider the view direction constraint satisfied.


When a new frame of point cloud is received, voxels of $\mathcal{M}_{obs}$ that intersected with the point cloud are identified and appended to a update queue $Q$. We subsequently iterate through the queue. For each voxel that is either unlabeled or previously labeled as poorly-observed, we will update its classification to well-observed if it satisfies both the distance and direction constraints. Otherwise, we label it as poorly-observed. {By updating only the voxels that intersect with the new point cloud frame, our method processes fewer voxels per update cycle, making it more computationally efficient than occupancy grid map based methods, which process all voxels within the LiDAR’s sensing range. }

\subsection{Frontier Detection and Clustering}
\label{subs:detection}

To achieve real-time frontier updates, we propose an incremental frontier detection and clustering method. We firstly iterate through all voxels that were labeled as poorly-observed in $Q$, and update its classification to frontier if the voxel has a neighboring voxel labeled as well-observed. Note that the frontier set $S_{ftr}$ is a subset of $S_{poorly}$. During the update process of $\mathcal{M}_{obs}$ described in Sec.\ref{subs:labelmap}, voxels with frontier labels are treated as poorly-observed voxels. 

During the iteration, an axis-aligned bounding box (AABB) $B_{update}$ of the voxels that changed from frontier to well-observed and those transitioned from poorly-observed to frontier are concurrently calculated.
In addition to updating $\mathcal{M}_{obs}$, newly generated frontiers in the current frame are also appended into a set $F_{current}$.

For the first frame of point clouds data, we apply a clustering approach to $F_{current}$ based on Euclidean distance and surface normals. Consider two frontier voxels $f_i$, $f_j$ with center positions $p_i$, $p_j$, and normal vectors $n_i$, $n_j$ respectively. We consider these two voxels to be neighbors if:
\begin{equation}
\|p_i - p_j\|_2 < \epsilon_d \quad \land \quad n_i \cdot n_j > \epsilon_n,
\end{equation}
where $\epsilon_d$ is a predefined distance threshold and $\epsilon_n$ is a normal similarity threshold.

The clustering process begins by selecting the first element from $F_{current}$.
Starting from the voxel, we employ a Breadth-First Search (BFS) algorithm to expand the cluster, 
identifying and incorporating the neighboring voxels that meet the specified distance and normal similarity criteria. 
The expansion terminates if the BFS is naturally completed or the size of the cluster's AABB exceeds a predefined threshold.
During this process, voxels added to clusters are simultaneously removed from $F_{current}$. 
Once the expansion of a cluster concludes, the process reinitializes with the first element of the updated $F_{current}$. The iteration continues until $F_{current}$ is empty. After that each cluster's  AABB, frontier voxels, and their corresponding normal vectors are stored in $\mathcal{L}_{clusters}$ for further processing.

For the second or subsequent frames of the point clouds, we firstly identify all clusters intersecting with $B_{update}$, remove them from $\mathcal{L}_{clusters}$ and add their frontier voxels into $F_{current}$. We then perform a new clustering process on $F_{current}$ and update  $\mathcal{L}_{clusters}$ based on the clustering results.

%% file: sections/planning.tex
\section{Hierarchical Exploration Planning}
\label{sec:path}
The hierarchical exploration planning module consists of three key components: an incrementally constructed topological graph $G$, a global guidance path planner, and a local trajectory planner.

To enable efficient path searching in large-scale environments, we propose an incremental topological graph constructed directly on point clouds. Leveraging the topological graph $G$, global planner computes guidance paths. Subsequently, the local planner generates safe, smooth, and feasible trajectories based on these guidance paths, followed by which the UAV can cover frontiers efficiently. 

\subsection{Incremental Topological Graph Construction}
\label{subs:topo_graph}
\begin{figure}[!t]
    \centering 
    \includegraphics[width=0.95\linewidth]{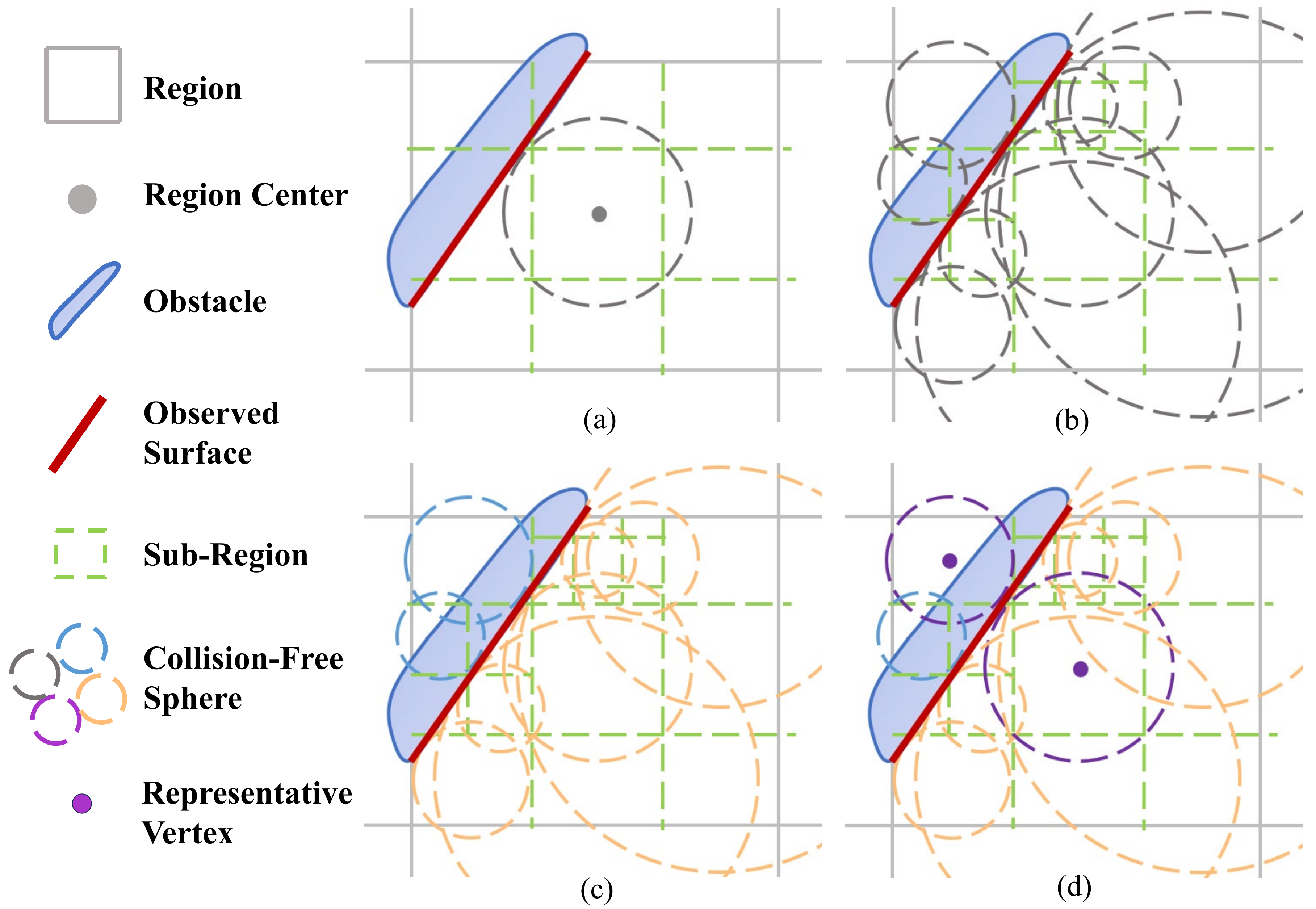}
    \caption{
   Illustration of topological graph vertex updating process in a region. (a) and (b): Free space coverage using collision-free spheres. (c): Clustering spheres based on connectivity. (d): Selection representative vertices.}
    \label{fig:region}
    \vspace{-0.7cm}
\end{figure}
Topological graph is widely employed in large-scale environment exploration to enable faster path searching\cite{zhang2024falcon, musil2022spheremap, yang2021graph}. However, most existing methods construct the topological graph based on a global occupancy grid map, which can be memory-intensive. To address this issue, we propose an incremental topological graph construction method that operates directly on point clouds. In our method, the topological graph is represented as $G = (V, E)$, where $V$ denotes the representative vertices in independent free spaces, and $E$ represents the edges showing connectivity between these free spaces. Each edge in $E$ is associated with a collision-free path.

Inspired by \cite{zhang2024falcon}, we decompose the entire exploration space into cuboid regions $\mathcal{R}$, which are managed by a spatial hash map. The vertices $V$ of the topological graph $G$ are stored within their corresponding regions.
Upon receiving a new point cloud frame, regions intersecting with the LiDAR's current observed areas are added to the update list $L_{region}$. Additionally, for regions added to $L_{region}$ for the first time throughout the exploration process, their corresponding vertex sets are initialized as $\emptyset$.

\subsubsection{Vertex Updating}
For each region $r_i$ in $L_{region}$, we employ multiple collision-free spheres to cover its free spaces. As illustrated in Fig. \ref{fig:region}(a) and Fig. \ref{fig:region}(b), the initial collision-free sphere is centered at $r_i$, with its radius determined by the distance to the nearest point in the point cloud map. If this sphere completely encompasses the region, the coverage operation for $r_i$ is concluded. If not, the process continues based on the sphere's radius. If the radius exceeds a predefined safety distance, the region is subdivided into smaller sub-regions by planes corresponding to the six faces of the sphere's inscribed cube. Otherwise, the sphere is discarded, the region is then split into eight sub-regions by three orthogonal planes through the sphere's center, parallel to the $xy$, $yz$, and $zx$ planes.

The process of sphere generation and region subdivision is recursively repeated for each sub-region until it's fully covered by collision-free spheres or its size is smaller than the safety radius. Subsequently, we employ a union-find algorithm\cite{unionset} to cluster these spheres based on their connectivity, as illustrated in Fig. \ref{fig:region}(c). Two spheres are considered part of the same cluster if their intersection size is larger than the safety radius. 

For each resulting cluster, a representative vertex is selected. The position of the vertex is set to the center of the sphere within the cluster that is closest to the region's center, as illustrated in Fig. \ref{fig:region}(d). Additionally, we record all vertices stored in $L_{region}$ as $V_{pre}$ before the process begins and as $V_{new}$ after $L_{region}$ is updated. Both sets are retained for subsequent algorithmic steps. 

\subsubsection{Edge Updating}
\begin{figure}[!t]
    \centering 
    \includegraphics[width=0.85\linewidth]{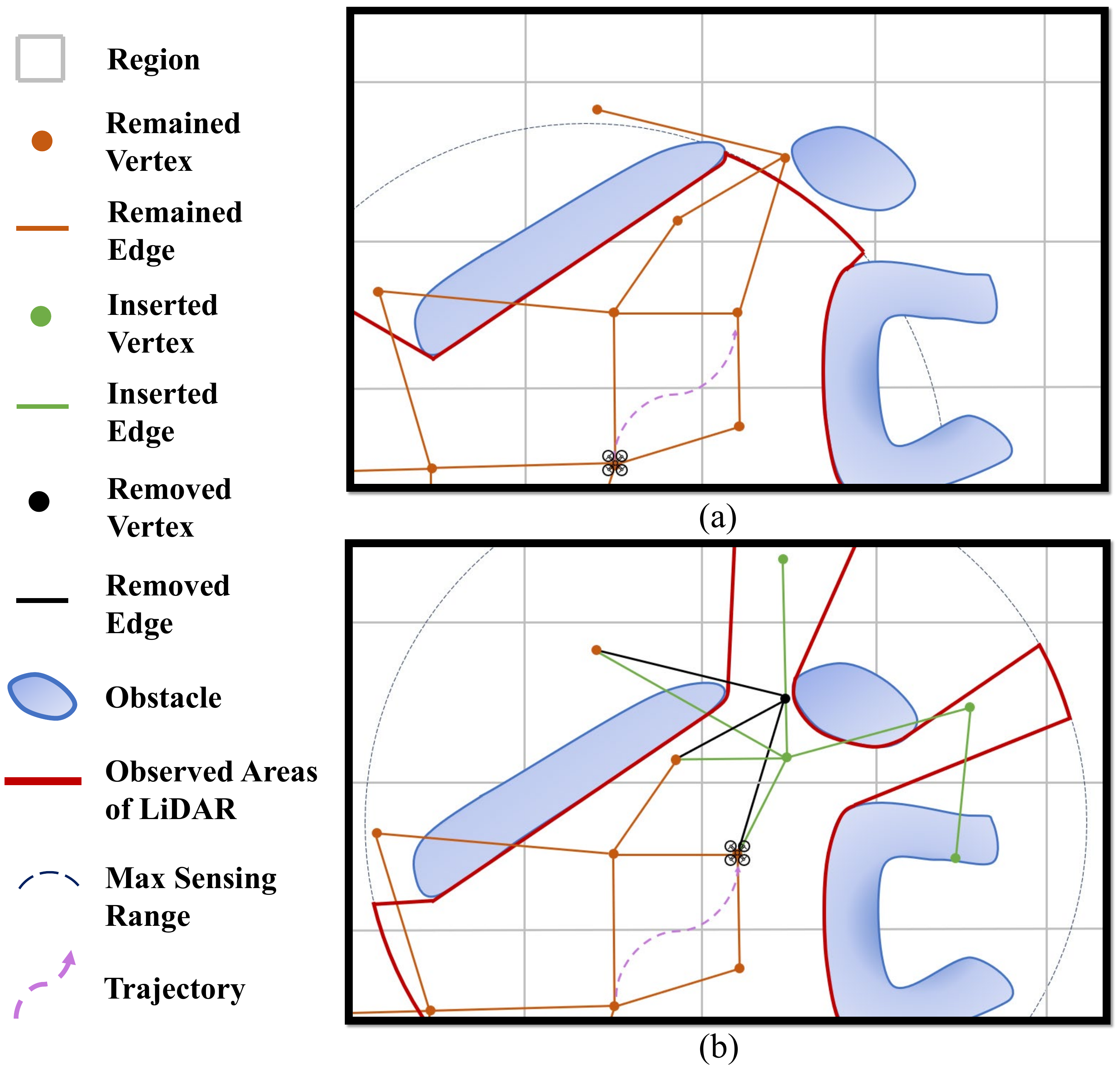}
    \caption{This figure shows the process of incremental updating of the topological graph as UAV flies along the trajectory. For clarity, we present a 2D example and assume that each region has only four neighbors (up, down, left, and right). However, in the actual implementation, to achieve better connectivity, we consider diagonally adjacent regions as neighbors as well.}
    \label{fig:topo}
    \vspace{-0.3cm}
\end{figure}
The edge set $E$ is initialized as $\emptyset$. As vertices of $G$ are updated, edges connecting them are derived from paths obtained through the A* search algorithm.
{As shown in Fig. \ref{fig:topo}(a), the initial topological graph is constructed at the starting position of the UAV, where vertices and edges are established based on the initial observations.} As the UAV flies along the trajectory, the topological graph is incrementally updated. Based on $V_{pre}$ and $V_{new}$ in newly observed areas, we categorize the vertices into three distinct sets:

\begin{equation}
\begin{aligned}
V_{remove} &= V_{pre} \setminus V_{new} \\
V_{remain} &= V_{new} \cap V_{pre} \\
V_{insert} &= V_{new} \setminus V_{pre} 
\end{aligned}
\end{equation}
where $V_{remove}$ represents the vertices that were present in $V_{pre}$ but are absent in $V_{new}$, $V_{remain}$ comprises the vertices that remain unchanged between $V_{pre}$ and $V_{new}$, $V_{insert}$ denotes the new vertices that appear in $V_{new}$ but were not in $V_{pre}$.

Firstly, we iterate through $V_{remove}$, eliminating all associated edges from $G$. Next, for vertices in $V_{remain}$, we examine all collision-free paths corresponding to edges connected to them. For each path, we verify whether it remains collision-free in the updated point cloud map. If a path is no longer collision-free, we add its corresponding vertex pair to a queue $Q_{edge}$ and remove the associated edge from $G$. Then, for each $v \in V_{insert}$, we construct vertex pairs between $v$ and nodes in neighboring regions, adding these pairs to $Q_{edge}$.

Subsequently, an A* search algorithm is performed for each vertex pair in $Q_{edge}$. If a collision-free path is found, the corresponding edge is then added to $G$, and the collision-free path is stored as well. Additionally, to accelerate the search process and avoid redundant edges, we constrain the search scope of the A* algorithm to a small area.  This area is defined as an axis-aligned bounding box (AABB) that tightly encloses the start and goal points, with an additional padding equal to the size of one region. 
Finally, we update $v_{odom}$, the vertex representing the UAV's current position, adjusting its position, removing previous edges, and establishing new connections. {The updated topological graph is shown in Fig. \ref{fig:topo}(b).}

\subsection{Global Guidance Path Planning}
\label{subsec:global_planning}
This module is designed to determine an efficient visitation sequence for frontier clusters in $\mathcal{L}_{clusters}$ and subsequently generate a guidance path for local trajectory planner. 

Given that globally optimal visitation sequences typically prioritize nearby clusters before distant ones, and considering the high-frequency replanning during flight, only a small portion of the guidance path is executed before the next replan cycle. 
Consequently the impact of distant cluster ordering is diminished. 
Therefore, we initially utilize a coarse distance estimation to rank all frontier clusters in $\mathcal{L}_{clusters}$, yielding a coarse preliminary visitation sequence. Then, we perform a more refined adjustment on the first K elements of this sequence, generating the global guidance path accordingly. This two-stage strategy concentrates computational resources on proximal clusters, effectively balancing path optimality with computational efficiency, making it particularly suitable for real-time exploration in large-scale scenarios.

\subsubsection{Frontier Clusters Sequencing}
During the exploration process, frontiers are continuously generated, each reachable via a straight-line path from the UAV's position at its generation time, as they are visible from the UAV at that moment. Inspired by this observation, we introduce a backtracking distance metric to prioritize frontier clusters. For a cluster centered at $c_i$, let $p_i$ denote the UAV's position at the time of the cluster's generation.
The backtracking distance $d_i$ is defined as:
\begin{equation}
d_i = d_{f}(p_i, p_{current}) + \|p_i - c_i\|_2, 
\end{equation}
where $d_{f}(p_i, p_{current}) $ represents the UAV's flight distance from $p_i$ to its current position, and $\|p_i - c_i\|_2$ denotes the euclidean distance between $p_i$ and $c_i$. 

The backtracking distance $d_i$ quantifies the total distance for the UAV to reach $c_i$ by retracing its historical path. By maintaining records of the UAV's flight distance at each cluster's generation time and its total flight distance, we can compute $d_i$ with $O(1)$ time complexity. This efficient computation allows us to rapidly rank clusters in $\mathcal{L}_{clusters}$, yielding a coarse global visitation sequence.

\subsubsection{Guidance Path Planning}

For the first K clusters in the sequence, we generate a viewpoint for each cluster based on the method described in Sec.\ref{sss:viewpoint}, and connect these viewpoints as vertices to graph $G$.
We then perform A* search on $G$ to compute distances between pairs of viewpoints. Subsequently, we formulate this as an Asymmetric Traveling Salesman Problem (ATSP) and solve it using \cite{lkh}. This solution yields the shortest path starting from the current vehicle position and traversing all selected viewpoints, serving as our global guidance path.

\subsubsection{Viewpoint Generation}
\label{sss:viewpoint}

\begin{figure}[!t]
    \centering 
    \vspace{-0.2cm}
    \includegraphics[width=0.95\linewidth]{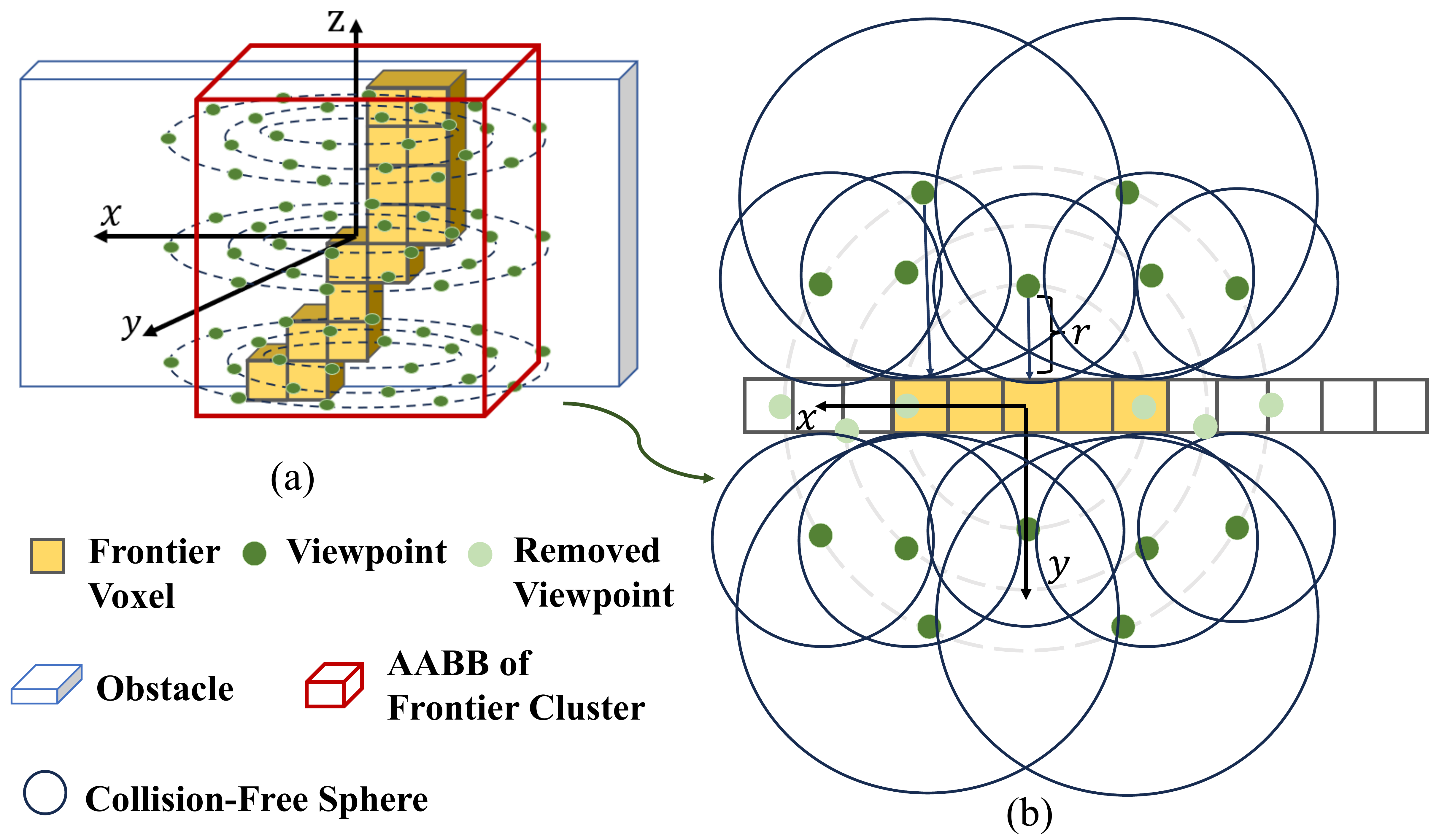}
    \caption{
   Viewpoint selection. (a): Uniformly sampling candidate viewpoints in a cylindrical coordinate system centered at the frontier cluster. (b): The process of clustering candidate viewpoints based on connectivity.}
    \label{fig:vp}
     \vspace{-2.0cm}
\end{figure}

For a frontier cluster $C_f$, candidate viewpoints are uniformly sampled in a cylindrical coordinate system centered at its center, as illustrated in Fig. \ref{fig:vp}(a). We then calculate the distance from each candidate viewpoint to the nearest point in the point cloud map. Viewpoints with distances below a safety threshold are eliminated. Remaining candidate viewpoints each correspond to a collision-free sphere. We then cluster them based on the connectivity of their associated spheres into several viewpoint clusters $C_{cv}$. This process is depicted in Fig. \ref{fig:vp}(b), where the candidate viewpoints are grouped into two viewpoint clusters.


For each cluster in $C_{cv}$, all elements inside are mutually accessible. So we select one representative element and add it to $G$ as a vertex $v_{represent}$ to check whether it's reachable from $v_{odom}$. If so, all elements within the cluster are considered reachable. After this assessment, $v_{represent}$ and its associated edges are removed from $G$ to ensure the graph remains efficient and clear for future planning cycles. We then select the viewpoint with the highest coverage score from the reachable candidates of $C_f$. 
The coverage score for a viewpoint is defined as the number of frontiers within $C_f$ that can be well observed from it. A frontier voxel is considered well-observed if the distance to the viewpoint is less than $D$, the line connecting the frontier voxel to the viewpoint $l_{vf}$ is unobstructed and the angle between $l_{vf}$ and the frontier's normal vector is below a threshold.
For each reachable viewpoint, candidate yaw angles are uniformly sampled over 360 degrees, and for each sampled yaw angle, the number of well-observed frontiers within the LiDAR’s field of view (FOV) is evaluated. The yaw angle that maximizes the number of well-observed frontiers is selected, and this count is recorded as the coverage score for the corresponding viewpoint. 
The viewpoint with the highest score is then selected and connected to the topological graph.

\vspace{-0.1cm}
\subsection{Local Trajectory Generation}
\label{subs:traj}
After obtaining the global guidance path, we generate smooth and energy-efficient local trajectories directly on the point cloud map. The piecewise polynomial MINCO\cite{wang2022geometrically} is adopted to separately represent position trajectory $\mathbf{p}(t) = [x(t), y(t), z(t)]^T$ and yaw trajectory $\psi(t)$. 
To construct the safety penalty, we generate a polyhedral safe flight corridor using FIRI\cite{FIRI} along the guidance path. 
To actively observe the unknown space, the yaw is constrained to look at future waypoints.
Additionally we apply constraints on maximum velocity, tilt angle, thrust, and body rates to ensure kinodynamic feasibility. A minimum time penalty is also incorporated to enable agile flight. Leveraging the properties of MINCO, we generate energy-efficient and agile local trajectories through joint spatio-temporal optimization.

%% file: sections/results.tex
\raggedbottom
\setlength{\tabcolsep}{4.4pt} 
\begin{table}[!t]
\centering
\caption{Results of Benchmark Comparisons}
\label{table:benchmark_com}
\renewcommand\arraystretch{1.1}
\begin{threeparttable} 

\begin{tabular}{c c c c c c c}
\toprule[0.8pt] 
\toprule[0.8pt] 
\textbf{Scene} & \textbf{Method} & \textbf{\makecell{Exp. \\ Time (s)}} & \textbf{\makecell{Avg. \\ Vel. (m/s) }}& \textbf{\makecell{Avg. \\ Tm. (s)}}& \textbf{\makecell{Max. \\ Tm. (s)}} & \textbf{\makecell{Exp. \\ Fin.}} \\ 
\hline
\multirow{4}[5]{*}{Cave}
                        & ERRT \cite{lindqvist2024tree} & -- & 2.19& 0.65 & 0.77  & No \\
                        & M3 \cite{gao2022meeting} & -- & 1.96 & \textbf{0.02} & \textbf{0.07} & No \\
                        & FUEL \cite{zhou2021fuel} & -- & 1.43 & 0.48 & 3.98 &No \\
                        & Proposed & \textbf{1288.2} & \textbf{2.68}& 0.09 & 0.18 & \textbf{Yes}\\
\hline
\multirow{4}[5]{*}{Garage}
                        & ERRT \cite{lindqvist2024tree} & -- & 1.92& 0.61 & 0.82 & No \\
                        & M3 \cite{gao2022meeting} & -- & 1.91& \textbf{0.02} &\textbf{0.06} & No \\
                        & FUEL \cite{zhou2021fuel} & 1486.5 & 1.55 & 0.48& 3.56 & \textbf{Yes}  \\
                        & Proposed & \textbf{716.9} & \textbf{3.10} & 0.05& 0.10 & \textbf{Yes} \\
\hline
\multirow{4}[5]{*}{Campus}
                        & ERRT \cite{lindqvist2024tree} & -- & 1.92 & 0.82 & 1.17 & No \\
                        & M3 \cite{gao2022meeting} & -- & 1.93 & \textbf{0.08} & 2.60 & No \\
                        & FUEL \cite{zhou2021fuel} & -- & 1.14 & 2.40 & 16.72 & No \\
                        & Proposed & \textbf{908.2} & \textbf{2.58} &\textbf{0.08}& \textbf{0.24} & \textbf{Yes}\\
\bottomrule[0.8pt] 
\bottomrule[0.8pt] 
\end{tabular}
\begin{tablenotes}
{\item[1] \footnotesize               
 From left to right, the columns represent: total exploration time, average velocity, average and maximum computation time per planning cycle, and whether the exploration is completed.
}
\end{tablenotes} 
\end{threeparttable} 
\vspace{-0.2cm}
\end{table}

\begin{table}[!t]
\centering
\caption{Results of Multi-Goal Path Searching}
\label{table:astar}
\renewcommand\arraystretch{1.2}
\begin{tabular}{c c c c}
\toprule[0.8pt] 
\toprule[0.8pt] 
\textbf{Planner} & PC-A* & OG-A* & TG-A* \\ 
\hline
{Total Time Consumption (ms)}& 190957.0  & 58255.2 & 12.2 \\
{Total Path Length (m) }& 1611.7  & 1568.7 & 1583.5 \\
\bottomrule[0.8pt] 
\bottomrule[0.8pt] 
\end{tabular}
\vspace{-0.2cm}
\end{table}

\begin{table}[!t]
\centering
\caption{Memory Consumption Comparisons}
\label{table:abl1}
\begin{threeparttable} 
\renewcommand\arraystretch{1.2}
\resizebox{1.0\linewidth}{!}{
\setlength{\tabcolsep}{2pt}
\begin{tabular}{c|cc|cc|cc|ccc}
\Xhline{1.2pt} 
\multicolumn{1}{c|}{\textbf{Trajectory}} & \multicolumn{2}{c|}{FUEL \cite{zhou2021fuel}} & \multicolumn{2}{c|}{ERRT \cite{lindqvist2024tree}} &  \multicolumn{2}{c|}{M3 \cite{gao2022meeting}} & \multicolumn{3}{c}{EPIC}\\
\cline{1-10} 
\diagbox[width=9em,height=5em]{\textbf{Scene}}{\textbf{MC$\dagger$}}{\textbf{Method}} & FUEL & EPIC & ERRT & EPIC & M3 & EPIC & ERRT & M3 & EPIC \\
\hline
Cave & 784.1 & \textbf{6.3} & 3.7 & \textbf{1.3} & 28.4 & \textbf{4.5} & 23.9 & 50.0 & \textbf{10.2} \\
\hline
Garage &32.1 & \textbf{5.2} & 11.3& \textbf{1.2} & 31.0 & \textbf{2.6} & 37.4 & 32.0 & \textbf{5.0} \\
\hline
Campus &207.5& \textbf{2.7} & 24.6& \textbf{2.4} & 16.5 & \textbf{2.0} & 47.4 & 30.0 & \textbf{4.9} \\
\Xhline{1.2pt} 
\end{tabular}
}
\vspace{0.1cm}
\begin{tablenotes}
\parbox[!]{0.46\textwidth}{\item[$\dagger$]\footnotesize              
Memory Consumption (MC) of the corresponding environment representation methods, measured in MB (megabytes).
}
\end{tablenotes}
\end{threeparttable} 
\vspace{-0.7cm}
\end{table}

\section{Experiments}
\label{sec:exp}
\subsection{Benchmark Comparisons}
\label{subs:benchmark}

\begin{figure*}[t]
	\centering
  \includegraphics[width= 2.0\columnwidth]{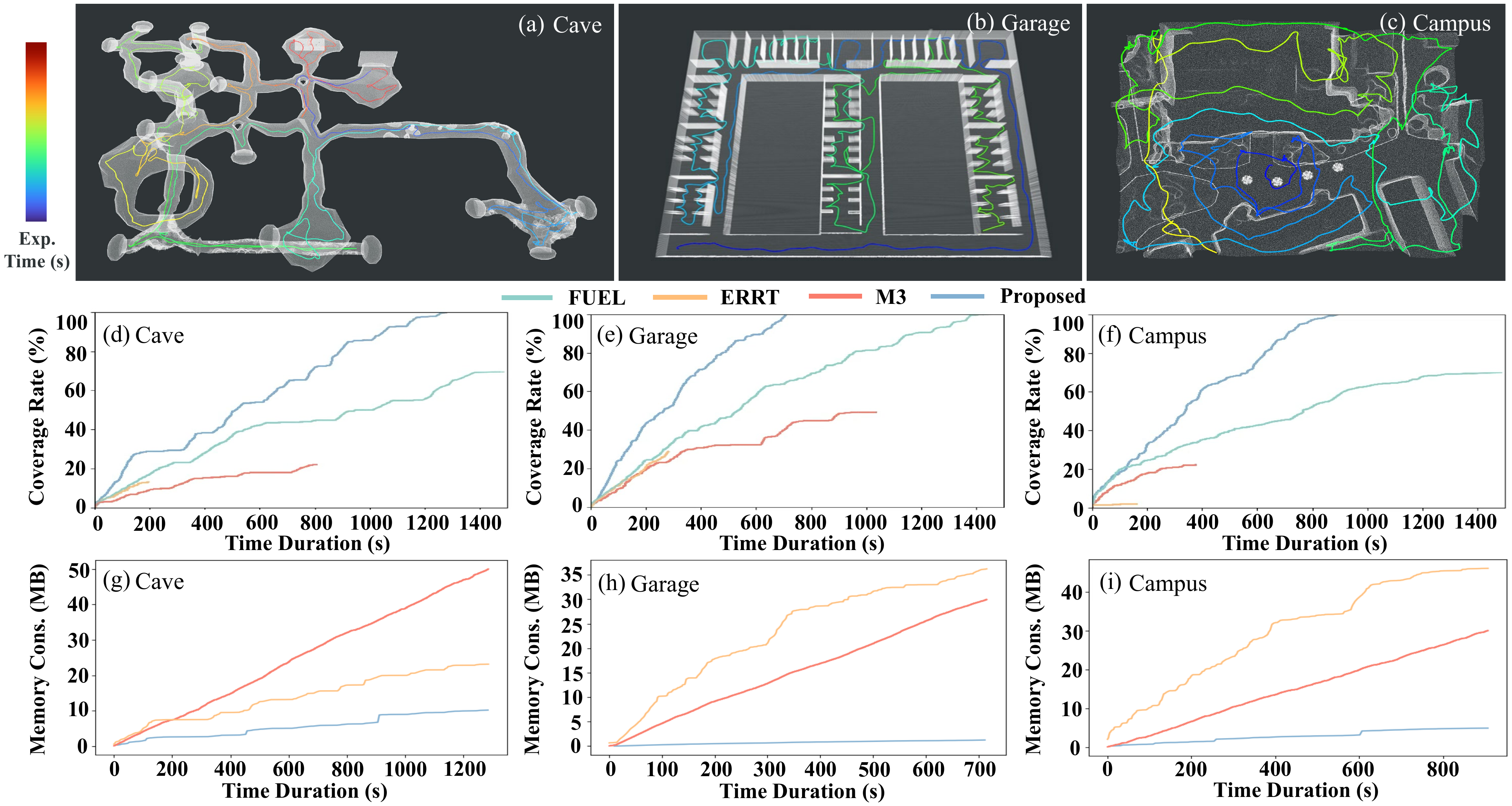} 
  \vspace{-0.3cm}
  \caption{\label{fig:mem_cons} (a)-(b) The trajectories of the proposed method in three scenes. (d)-(f) The exploration progress of all the three state-of-the-arts benchmarks and the proposed method in three scenes. { (g)-(i) Memory Consumption (Memory Cons.) over time for ERRT, M3, and the proposed method in three scenes, with all methods following the proposed method's trajectory. FUEL is excluded as its pre-allocated global uniform occupancy grid map results in constant, high memory usage, making it less informative for this comparison.}}
  \vspace{-0.5cm}
\end{figure*}

In this section, we evaluate our method against three state-of-the-art exploration algorithms: FUEL \cite{zhou2021fuel}, ERRT \cite{lindqvist2024tree}, and M3 \cite{gao2022meeting} using the MARSIM simulator \cite{kong2023marsim}. {These methods represent three distinct exploration frameworks with different environment representations. As state-of-the-art exploration techniques, they provide a fair and reproducible basis for comparison.} Tests were conducted in three large-scale scenes: a cave {$[224\times330\times41] \  m^3$}, a garage {$[192\times156\times4]\ m^3$}, and a campus {$[160\times128\times40]\ m^3$}. {The cave scene is based on the DARPA SubT Cave World dataset described in \cite{akoval2020}.} Each method was executed four times per scene with a maximum velocity of {$4.0\ m/s$}, map resolution of {$0.4\ m$} (except M3), and a 1500-second time limit. All tests were performed on an Intel Core i5-13600KF CPU with 32 GB RAM. 
Table \ref{table:benchmark_com} provides performance statistics, while Fig. \ref{fig:mem_cons} (a)-(f) show the trajectories of our method and the exploration progress of all four methods in three scenes, respectively.

In the cave scene, characterized by vertical drops, narrow tunnels, and dead ends, as well as in the campus scene, featured varying building heights and extensive open areas, our method was the only one to achieve complete exploration. FUEL timed out due to the increasing computational burden of path searching as the explored area expand, ERRT terminated early due to insufficient information gain, and M3 prematurely ended its exploration due to its coarse frontier update strategy. 
In the garage scene, a more structured environment, both our method and FUEL completed the exploration, though FUEL required twice as much time. ERRT and M3 achieved relatively high surface coverage but did not complete the task.

Each method exhibits distinct trade-offs: M3 and ERRT both employ greedy strategies for stable response times but sacrifice global planning optimality. Note that although M3 achieves the lowest average computation time in most scenes, it fails to complete the exploration due to its overly simplified representation. FUEL, on the other hand, prioritizes global planning optimality, leading to high computational burden (up to {$16.72\ s$} in the campus scene).
Our algorithm balances these approaches, with competitive computation time ({$0.05-0.09\ s$} on average),  achieving complete exploration and near-optimal path planning (even shorter than FUEL in the garage scene). This results in higher average velocities  ({$2.5-3.1\ m/s$} on average) and faster overall exploration.



\vspace{-0.2cm}
\subsection{Ablation Experiments}
\label{subs:Ablation}




\emph{1) The Efficiency of Path Searching:} A* path searching is frequently executed by the guidance path planning module, significantly impacting overall system performance.
To demonstrate the importance of topological graphs in accelerating searches, 
we conducted a comparative study of three A* implementations: our Topological Graph-based A* (TG-A*), Point Cloud A* (PC-A*) which applies A* directly on point cloud map, performing collision checks using Nearest Neighbor Search (NNS) algorithm, and Occupancy Grid A* (OG-A*), the most common A* implementation, which performs collision detection based on querying occupancy information. 
We tested these methods in a simulated garage environment with 12 goal points from a fixed starting position. 
Table \ref{table:astar} presents total search time and path length.
Our TG-A* approach maintains comparable path quality while achieving speed improvements of 3 to 4 orders of magnitude, significantly enhancing exploration efficiency.


\emph{2) Evaluation of Memory Consumption:} {We evaluate the memory consumption of each method by measuring its key environment representation: FUEL \cite{zhou2021fuel}’s global uniform occupancy grid map, ERRT \cite{lindqvist2024tree}’s UFOMap \cite{duberg2020ufomap}, M3 \cite{gao2022meeting}’s star-convex polytopes \cite{zhong2020generating} generated every 1.5 meters of flight, and EPIC’s observation map. These structures form the foundation for frontier extraction and exploration guidance in each method. To ensure a fair and comprehensive comparison, we conducted two sets of experiments. First, each method explored the three scenes independently, using its own strategy. Second, we cross-evaluated the methods by having EPIC execute the trajectories generated by the other three methods, and vice versa, with the other methods running EPIC’s trajectory. The results consistently demonstrate that our environment representation achieves the lowest memory consumption across all tested scenarios
as shown in Table \ref{table:abl1} and Fig. \ref{fig:mem_cons} (g)-(i).}


\vspace{-0.2cm}
\subsection{Real-World Experiments}
\label{subs:realworld}
\begin{figure}[!t]
    \centering 
    \vspace{-0.4cm}
    \includegraphics[width=0.8\linewidth]{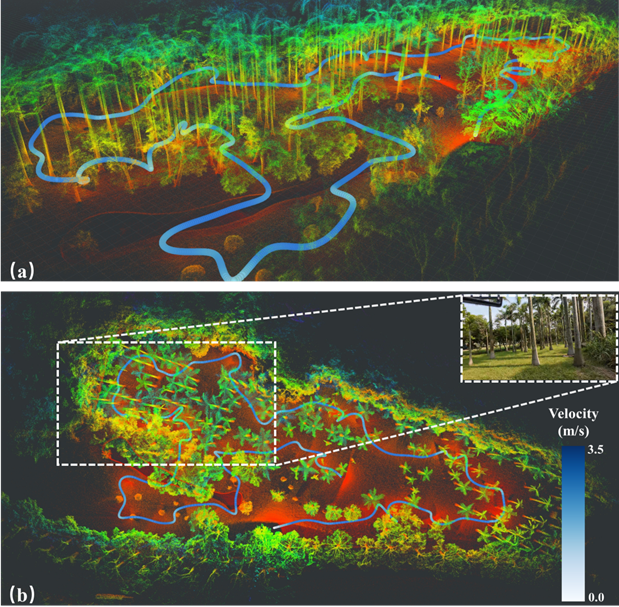}
    \caption{
    Performing exploration task in a forest environment. (a) Side view of the online-built point cloud map with the UAV’s executed trajectory. (b) Top-down view of the same point cloud map. 
    }
    \label{fig:forest}
    \vspace{-0.9cm}
\end{figure}

Various real-world experiments were conducted to further validate our method. We utilized a LiDAR-based quadrotor platform equipped with an Intel NUC onboard computer, a NxtPx4 autopilot\cite{liu2024omninxt}, and a Livox MID360 LiDAR mounted at a $40^\circ$ pitch angle. A modified version of FAST-LIO2\cite{xu2022fast} was adapted for localization and mapping. The maximum speed  was limited to {$3.5\ m/s$} to ensure safety.
First, we explored a large-scale campus building scene. The exploration covered a {$[77\times74\times3]\ m^3$} area, taking 181 seconds and traversing {$530.4\ m$}. A sample point cloud map and flight trajectory are presented in { Fig. 1.}
Second, to validate our method in natural environments, we conducted exploration tests in a {$[50 \times 84 \times 3] \ m^3$} cluttered forest scene, taking 152.6 seconds and traversing {$328.5\ m$}. The point cloud map and trajectory are displayed in Fig. \ref{fig:forest}.
Finally, we tested our method in a {$[77\times24\times3] \ m^3$} indoor corridor scene, taking 94.7 seconds with a {$194.8\ m$} flight length. More detailed experimental results and visualizations of all scenarios are available in the supplementary video.


    